\documentclass{article}
\usepackage{spconf,amsmath,graphicx}

\usepackage{times}  
\usepackage{helvet}  
\usepackage{courier}  
\usepackage{url}  
\usepackage{graphicx}  
\usepackage{multirow}
\usepackage{amsmath}

\usepackage{CJK}
\usepackage{bm}

\title{Condition-Transforming Variational AutoEncoder for Conversation Response Generation}
\name{Yu-Ping Ruan$^1$, Zhen-Hua Ling$^1$, Quan Liu$^2$, Zhigang Chen$^2$, Nitin Indurkhya$^1$\thanks{This work was partially funded by the National
Nature Science Foundation of China (Grant No. U1636201).}}
\address{$^1$National Engineering Laboratory for Speech and Language Information Processing,\\
University of Science and Technology of China, Hefei, P.R.China\\
$^2$iFLYTEK Research, Hefei, P.R. China\\
{\tt \small  ypruan@mail.ustc.edu.cn, \{zhling, nitin\}@ustc.edu.cn, \{quanliu, zgchen\}@iflytek.com}\\
}
\begin{document}
\ninept
\begin{CJK}{UTF8}{gbsn}
\maketitle
\begin{abstract}
  This paper proposes a new model, called condition-transforming variational autoencoder (CTVAE), to improve the performance of conversation response generation using conditional variational autoencoders (CVAEs). In conventional CVAEs , the prior distribution of latent variable z follows a multivariate Gaussian distribution with mean and variance modulated by the input conditions. Previous
work found that this distribution tends to become condition-independent in practical application. In our proposed
  CTVAE model, the latent  variable $\mathbf{z}$ is sampled by performing a non-linear transformation on the combination of the input conditions and the samples from a condition-independent prior distribution $\mathcal{N}(\mathbf{0}, \mathbf{I})$.
  In our objective evaluations, the CTVAE model outperforms the CVAE model on fluency metrics and surpasses a sequence-to-sequence (Seq2Seq) model on diversity metrics.
  In subjective preference tests, our proposed CTVAE model performs significantly better than CVAE and Seq2Seq models on generating fluency, informative and topic relevant responses.
\end{abstract}
\begin{keywords}
variational, autoencoders, conversation, text generation
\end{keywords}
\section{Introduction}
\label{introduction}
There has been a growing interest in neural-network-based end-to-end models for text generation tasks, including machine translation \cite{bahdanau2014neural}, text summarization \cite{DBLP:conf/emnlp/RushCW15}, and conversation response generation \cite{DBLP:conf/nips/SutskeverVL14,DBLP:conf/acl/ShangLL15,DBLP:conf/aaai/SerbanSBCP16}. Among these, encoder-decoder framework has been widely adopted and they principally learn the mapping from an input sequence $x$ to its target sequence $y$.
Although this framework has achieved great success in machine translation,
previous studies on generating responses for chit-chat conversations \cite{DBLP:conf/aaai/SerbanSBCP16,li2015diversity}
have found that ordinary encoder-decoder models tend to generate dull, repeated and generic responses in conversations,  such as \emph{``i don't know", ``that's ok"}, which are lack of diversity.
One possible reason is the deterministic calculation of ordinary encoder-decoder models which constrains them
from learning the $1$-to-$n$ mapping relationship, especially on semantic connections, between input sequence and potential multiple target sequences.
In the task of chit-chat conversation, modeling and generating the diversity of responses is important because
an input post or context may correspond to multiple responses with different meanings and language styles.

Many attempts have been made to alleviate these deficiencies of encoder-decoder models, such as by utilizing extra features or knowledge as conditions to generate more specific responses \cite{xing2017topic,ghazvininejad2017knowledge} 
and by improving the model structure, the training algorithms and the decoding strategies \cite{wu2017neural,li2016deep,zhou2017mechanism}.
Additionally, conditional variational autoencoders (CVAEs),  which were originally proposed for image generation \cite{sohn2015learning,yan2016attribute2image},
have recently been applied to dialog response generation \cite{zhao2017learning,serban2017hierarchical}.
Variational generative models, including variational autoencoders (VAEs) and CVAEs, are suitable for learning the $1$-to-$n$ mapping relationship due to their variational sampling mechanism for deriving latent representations.

This paper studies variational generative models for text generation in single-turn chit-chat conversations.
The CVAE models used in previous  work \cite{zhao2017learning,yang2017generating,sohn2015learning,yan2016attribute2image,serban2017hierarchical} all assumed a prior distribution of latent variable $\mathbf{z}$  followed a multivariate Gaussian distribution $p_\theta(\mathbf{z}|x)$
whose mean and variance were estimated by a prior network using condition $x$ as input.
However, previous studies on image generation \cite{sohn2015learning,kingma2014semi} found that the samples of $\mathbf{z}$ from $p_\theta(\mathbf{z}|x)$ tended to be independent of $x$ given estimated models,
which implied that the effect of the condition $x$ was constrained at the generation stage.
In the conversation response generation task, the condition $x$ is in the form of natural language.
The semantic space of $x$  in the training set is always sparse,
which further increases the difficulty of estimating  the prior network $p_\theta(\mathbf{z}|x)$.
To address this issue of CVAEs, we propose condition-transforming variational autoencoders (CTVAEs) in this paper.
In contrast to CVAEs, which use prior networks to describe $p_\theta(\mathbf{z}|x)$,
a condition-independent prior distribution $\mathcal{N}(\mathbf{0}, \mathbf{I})$ is adopted in CTVAEs.
Then, another transformation network is built to  derive the samples of $\mathbf{z}$ for decoding
by transforming the combination of condition $x$ and samples from $\mathcal{N}(\mathbf{0}, \mathbf{I})$.

Specifically, the contributions of this paper are two-fold: First, the subjective preference tests in this paper demonstrate that there is no significant performance gap between
the ordinary CVAE model and a simplified CVAE model whose prior distribution is fixed as a condition-independent distribution, i.e., $\mathcal{N}(\mathbf{0}, \mathbf{I})$, which implies that the effects of the condition-dependent prior distribution in CVAE were limited.
Second, a new model, called CTVAE, is proposed to enhance the effect of the conditions in CVAEs.
This model samples the condition-dependent latent variable $\mathbf{z}$  by
performing a non-linear transformation on the combination of the input condition and the samples from a condition-independent Gaussian distribution.
In our experiments of generating short text conversations, the CTVAE model outperforms CVAE on objective fluency metrics and surpasses a sequence-to-sequence (Seq2Seq) model on objective diversity metrics. In subjective preference tests, our proposed CTVAE model performs significantly better than CVAE and Seq2Seq models on generating fluency, informative and topic relevant responses.

\section{Methodology}
\subsection{From CVAE to CTVAE}
\begin{figure}[!t]
	\centering
	\includegraphics[width=3.4in]{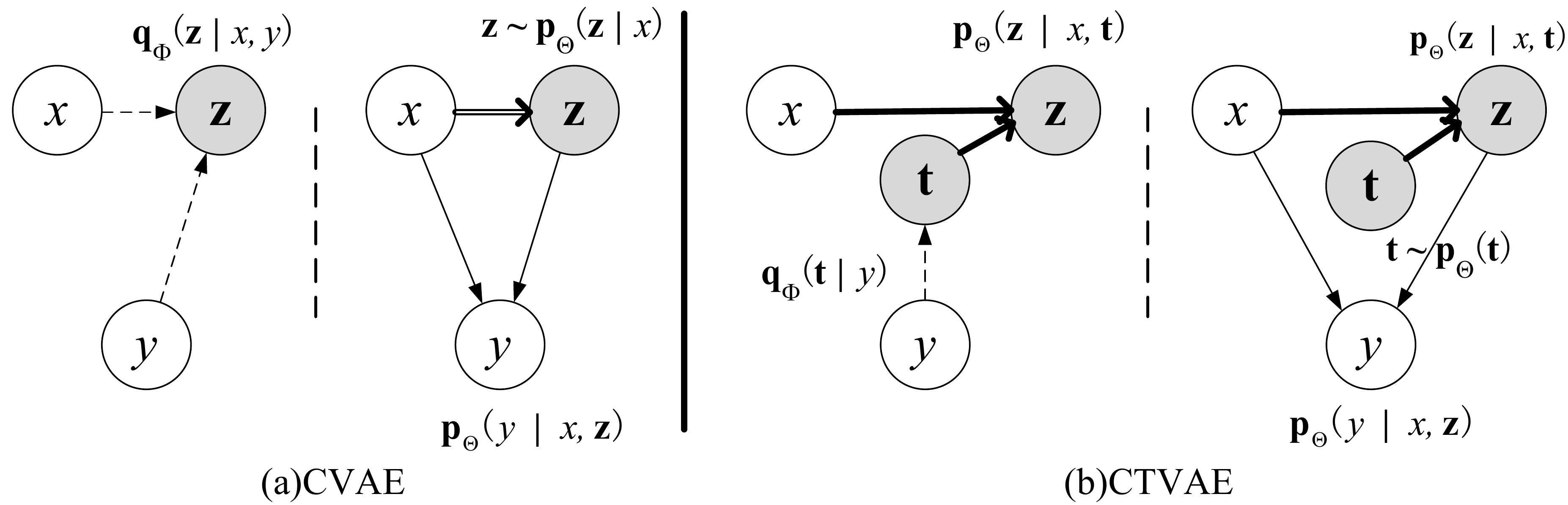}
	\caption{Graphical models of (a) CVAE, and (b) CTVAE. In each subgraph, the left part shows the recognition process of latent variable $\mathbf{z}$  during the training stage, and the right part shows the process of generating $y$ during the testing stage.
The dashed lines and the single solid lines represent the recognition network and the decoder network respectively.
The double solid line in (a) and the thick solid lines in (b) denote the prior network and the transformation network respectively.
}\label{fig:graphical}
\end{figure}

\begin{figure*}[!t]
	\centering
	\includegraphics[width=5.2in]{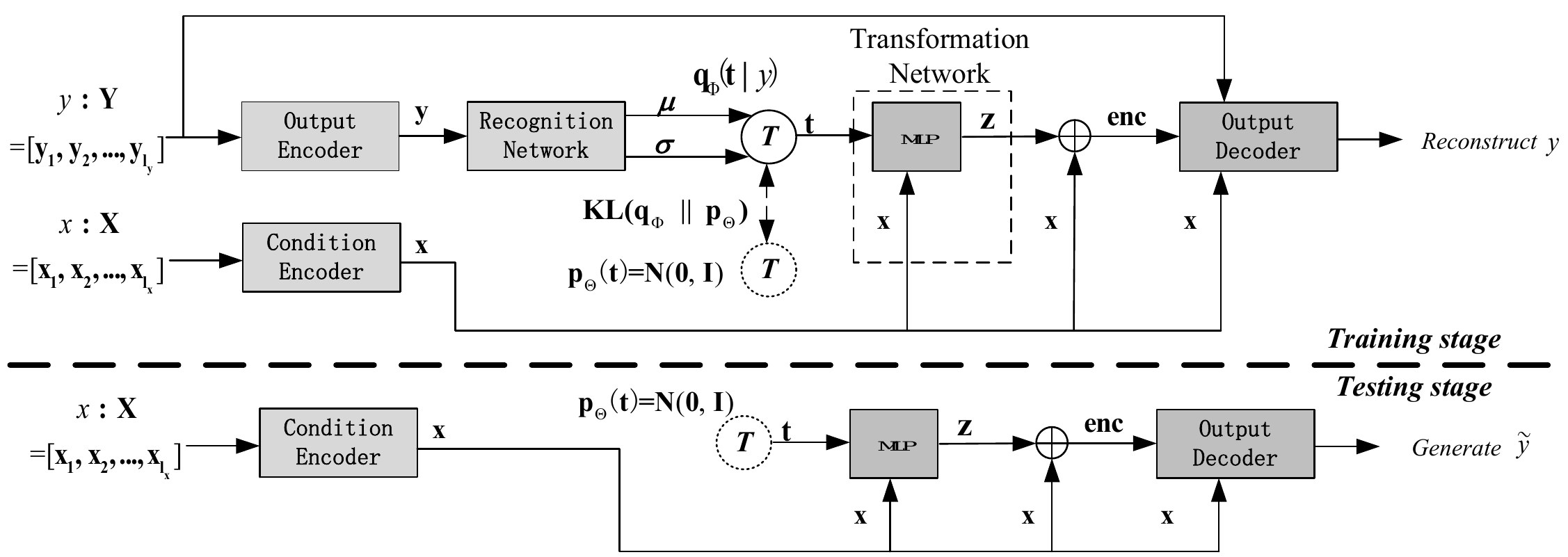}
	\caption{The model architecture of the CTVAE implemented in this paper. $\bigoplus$ denotes the concatenation of input vectors. All the encoders and decoders are 1-layer LSTM-RNNs, both recognition network and transformation network are MLPs.}
	\label{fig:models}
\end{figure*}
Figure \ref{fig:graphical} shows directed graphical models of CVAE and CTVAE.
In the single-turn short text conversation task,
the condition $x$ is the input post and ${y}$ is the output response.
As Figure \ref{fig:graphical}(a) shows, a CVAE is composed of a prior network $p_\theta(\mathbf{z}|x)$, a recognition network $q_\phi(\mathbf{z}|x,y)$, and a decoder network $p_\theta(y|x,\mathbf{z})$. Both $p_\theta(\mathbf{z}|x)$ and $q_\phi(\mathbf{z}|x,y)$ are multivariate Gaussian distributions. The generative process of response $y$ at testing stage is as follows: sample a $\mathbf{z}$ point from the prior distribution $p_\theta(\mathbf{z}|x)$, then feed it into decoder network $p_\theta(y|x,\mathbf{z})$.
CVAEs can be efficiently trained with the stochastic gradient variational Bayes (SGVB) \cite{kingma2013auto} framework by maximizing the lower bound of the conditional log likelihood $\log p(y|x)$ as follows,
\begin{equation}\label{eq:cvae_loss}
\begin{aligned}
\mathcal{L}_{CVAE}&(\theta,\phi;x,y)=-KL(q_\phi(\mathbf{z}|x,y)||p_\theta(\mathbf{z}|x))\\
&+\mathbf{E}_{q_\phi(\mathbf{z}|x,y)}[\log p_\theta(y|x,\mathbf{z})]\le\log p(y|x).
\end{aligned}
\end{equation}
As shown in Figure \ref{fig:graphical}(b), a CTVAE has no prior network but adopts  $\mathcal{N}(\mathbf{0}, \mathbf{I})$ as a transitional prior distribution $p_\theta(\mathbf{t})$ to generate $y$.
Similarly, a CTVAE includes a recognition network $q_\phi(\mathbf{t}|y)$ and a decoder network $p_\theta(y|x,\mathbf{z})$.
Additionally, CTVAEs use an alternative non-linear transformation network $p_\theta(\mathbf{z}|x,\mathbf{t})$ to sample the latent variable $\mathbf{z}$ from the combination of $x$ and the samples of transitional latent variable $\mathbf{t}$.
Following the training strategy for CVAEs,
the model parameters of CTVAEs can be estimated by maximizing the lower bound of the conditional log likelihood $\log p(y|x)$ as follows,
\begin{equation}\label{eq:CTVAE_loss}
\begin{aligned}
&\mathcal{L}_{CTVAE}(\theta,\phi;x,y)=-KL(q_\phi(\mathbf{t}|y)||p_\theta\mathbf{(t)})\\
&+\mathbf{E}_{p_\theta(\mathbf{z}|x,\mathbf{t})q_\phi(\mathbf{t}|y)}[\log p_\theta(y|x,\mathbf{z})]\le \log p(y|x).
\end{aligned}
\end{equation}
\subsection{Model Implementation} \label{model_implementation}

The model architecture of the CTVAE implemented in this paper is shown in Figure \ref{fig:models}.
Specifically, all the \emph{encoders} and \emph{decoders} are 1-layer recurrent neural networks with long short-term memory units.
For an input post $x=[x_1, x_2, ...,x_{l_x}]$ with $l_x$ words, we can derive the corresponding output hidden states $[\mathbf{h}_1, \mathbf{h}_2, ...,\mathbf{h}_{l_x}]$ by sending its word embedding sequence $\mathbf{X}=[\mathbf{x}_1, \mathbf{x}_2, ...,\mathbf{x}_{l_x}]$ into the \emph{Condition Encoder}.
Then, the mean pooling of hidden states $[\mathbf{h}_1, \mathbf{h}_2, ...,\mathbf{h}_{l_x}]$ is used to present the condition post, denoted as $\mathbf{x}$. Similarly, we can derive vector representation $\mathbf{y}$ for response $y$ by inputting $\mathbf{Y}=[\mathbf{y}_1, \mathbf{y}_2, ...,\mathbf{y}_{l_y}]$ into the \emph{Output Encoder}.
The \emph{Recognition Network} is a multi-layer perceptron (MLP), which has a hidden layer with \emph{softplus} activation and a linear output layer in our implementation.
The recognition network predicts $\mathbf{\bm{\mu}}$ and $\log(\mathbf{\bm{\sigma}}^2)$ from $\mathbf{y}$, which gives $q_\phi(\mathbf{t}|y)=\mathcal{N}\mathbf{(\bm{\mu}, {\bm{\sigma}}^2I)}$.
The samples of the transitional latent variable $\mathbf{t}$ generated by $q_\phi(\mathbf{t}|y)$ are further used to derive the samples of latent variable $\mathbf{z}$ for reconstructing $y$ during training.
To guarantee the feasibility of error backpropagation for model training,
reparametrization \cite{kingma2013auto} is performed to generate the samples of $\mathbf{t}$.
To derive the samples of latent variable $\mathbf{z}$, the sampled $\mathbf{t}$ is concatenated with condition $\mathbf{x}$ and passed through a transformation network, which is a MLP with two hidden layers with \emph{tanh} activation in our implementation. The output of the transformation network is used as the samples of latent variable $\mathbf{z}$.
The initial hidden state of the $Output Decoder$  is $\mathbf{x}$.
At each time step of the 1-layer LSTM-RNN, the input is composed of the word embedding from the previous time step and the encoding vector $\mathbf{enc}$, which is the concatenation of $\mathbf{x}$ and $\mathbf{z}$ samples.
According to Eq. (\ref{eq:CTVAE_loss}), the summation of the log-likelihood of reconstructing $y$ from the $Output Decoder$ and the negative KL divergence between $q_\phi(\mathbf{t}|y)$
and the prior distribution  of the transitional latent variable $p_\theta(\mathbf{t})=\mathcal{N}(\mathbf{0}, \mathbf{I})$
is used as the objective function for  training.

In the CVAE built for comparison, all its encoders and decoder have identical structure to those in  CTVAE. Both the recognition network and prior
network have the same structure as the recognition network in CTVAE except that the recognition network accepts the concatenation of $\mathbf{x}$ and $\mathbf{y}$ as input.

\subsection{Reranking Multiple Responses}\label{resp_rank}
To evaluate the performance of producing diverse responses using different models, multiple responses for each post are generated at the testing stage.
Specifically, for the CVAE/CTVAE models, we first generatd multiple samples of $\mathbf{z}$.
Then, for each $\mathbf{z}$ sample, a beam search is adopted to return the best result.
The multiple responses for each post are reranked using a topic coherence discrimination (TCD) model, which is trained based on the ESIM model \cite{chen2017enhanced}.
Specifically, we replace all BiLSTMs in the ESIM with 1-layer LSTMs and define the objective of the TCD model as judging whether a response is a valid response to a given post.
In order to train the TCD model, all post-response pairs in the training set are used as positive samples and negative samples are constructed by randomly shuffling the mapping between posts and responses. Finally, ranking scores are adopted to rerank all responses generated for one post.
The scores are calculated as $\log p_\theta(\tilde{y}|c) + \lambda * \log{p_{TCD}(true|x,\tilde{y})}$,
where the first term is the log-likelihood of generating response $\widetilde{y}$ using the decoder network and $c$ is the condition input to the decoder, i.e., $[x,\mathbf{z}]$ in CVAE/CTVAE models. The second term is the log-likelihood of the output probability of the TCD model.
$\lambda$ represents the weight between  these two terms.

\section{Experiments}
\subsection{Dataset}
The short text conversation (STC) dataset from \emph{NTCIR-12} \cite{shang2016overview} was used in our experiments.
This dataset was crawled from Chinese Sina Weibo\footnote{\url{https://weibo.com/}}.
The dataset contains $1,800,000$ post-response pairs\footnote{This dataset was originally prepared for retrieval models and had no standard division for generative models. Here we filtered the post-response pairs in raw STC dataset according to word frequencies to built our dataset.}, and one post corresponds to an average of $19$ responses.
Therefore, it contains $1$-to-$n$ mapping relationship and is appropriate for studying diverse text generation methods.
We randomly split the data into $1,708,415/73,120/18,465$ pairs to build the training, development and test sets.
There were no overlapping posts among these three sets.

\subsection{Models in Our Experiments}
In our experiments, we compared \emph{CTVAE} with following three baseline models, i.e., \emph{Seq2Seq, CVAE-simple}, and \emph{CVAE} . We didn't include the models in \emph{NTCIR-12} contest because they were all retrieval models.

$\bullet$ \textbf{Seq2Seq} Like previous study, we used the encoder-decoder neural network with attention as the baseline model  \cite{zhao2017learning,serban2017hierarchical}, which was similar to that for machine translation \cite{bahdanau2014neural,luong2015effective}. Both the encoder and decoder were 1-layer LSTM-RNNs, and the attention weights were obtained by the inner product of the hidden states.

$\bullet$ \textbf{CVAE-simple \& CVAE} The CVAE model has been described in Section \ref{model_implementation}. As described in Section \ref{introduction}, the prior distribution $p_\theta(\mathbf{z}|x)$ in CVAEs was previously found to degrade to $p_\theta\mathbf{(z)}$. To verify this, we manually removed the prior network $p_\theta(\mathbf{z}|x)$ in the CVAE and fixed the prior distribution to $p_\theta\mathbf{(z)}=\mathcal{N}\mathbf{(0,I)}$.
This modified  CVAE model was denoted as CVAE-simple.

\subsection{Parameter Setting}\label{traing_details}
We trained the models in our experiments with the following hyperparameters.
All word embeddings, hidden layers of the recognition network and prior network, hidden layers of the transformation network, and hidden state vectors of the encoders and decoders had $300$ dimensions. The latent variables $\mathbf{t}$ in CTVAE and $\mathbf{z}$ in CVAE had $100$ dimensions.
Each encoder and decoder had word embeddings of its own, and the vocabulary size was $35,000$. All word embeddings and model parameters were initialized randomly with Gaussian-distributed samples. The method of Adam \cite{kingma2014adam} was adopted for optimization with initial learning rate $5e-04$. The batch size was set to $128$.
When training the CVAEs and CTVAEs,
the \emph{KL annealing} strategy \cite{bowman2016generating} was adopted to address the issue of  latent variable vanishing.
The model parameters were pre-trained without optimizing the KL divergence term.
Additionally, we also adopted a training strategy which optimized the KLD loss term every 3 steps but optimized the reconstruction non-negative log likelihood (NLL) loss term every 1 step.
As described in Section \ref{resp_rank}, we generated multiple responses for each post. Specifically, for CVAE and CTVAE models, the number of $\mathbf{z}$ samples was set to 50, the beam search size was 20. For Seq2Seq, a beam search with beam size 50 was used to return multiple responses.
The weight $\lambda$ for reranking was heuristically set to $5$.
The top-5 responses after reranking were used for evaluation in our experiments.

\subsection{Objective Evaluation}

\begin{table}[!t]
	\centering
	\begin{tabular}{l|cccc}
		\hline
		& Seq2Seq & CVAE-simple & CVAE & CTVAE\\
		\hline
        PPL on LM & 7.61 & 31.82 & 36.96 & 21.75\\
        Matching(\%) & 92.58 & 8.12 & 10.51 & 19.10\\
        \hline
	\end{tabular}
    \caption{The objective \emph{fluency} performance of different models.}
	\label{tab:object_fluency}
\end{table}

\begin{table}[!t]
	\centering
	\begin{tabular}{l|cccc}
		\hline
		& Seq2Seq & CVAE-simple & CVAE & CTVAE\\
		\hline
        Distinct-1(\%) & 1.61 & 10.26 & 11.52 & 8.69\\
        Distinct-2(\%) & 5.26 & 41.23 & 42.6 & 33.44\\
        Unique(\%) & 22.86 & 97.66 & 97.78 & 97.62\\
        \hline
	\end{tabular}
    \caption{The objective \emph{diversity} performance of different models.}
	\label{tab:object_diversity}
\end{table}

\subsubsection{Fluency}\label{fluency_results}
We trained a RNN language model (LM) \cite{mikolov2010recurrent} using the same STC dataset to evaluate the fluency of the generated responses by calculating their perplexities, denoted as \emph{PPL on LM} here.
Furthermore, the percentage of generated responses that exactly matched any responses in the training set were counted.
This matching percentage was used as a metric to evaluate the model's ability to generate fluency sentences with reasonable syntactic and semantic representations.
For each model, 50 responses were generated for $512$ unique posts in the test set, and the responses were reranked using the methods described in Section \ref{resp_rank}.
The average LM perplexity and matching percentage of all top-5 responses were calculated for each model and the results are presented in Table \ref{tab:object_fluency}.
It can be found that the Seq2Seq model
achieved the lowest perplexity on LM and the highest matching percentage because it tended to generate its dull, generic and repeated responses.
The CVAE models performed worst on these two fluency metrics.
For CTVAE, it performed much better than the CVAE models on both LM perplexity and matching percentage.
\begin{table*}[!t]

    \small
	\centering
	\begin{tabular}{c|c|ccccc|c}
		\hline
		& \textbf{Pair No.} & \textbf{Seq2Seq} & \textbf{CVAE-simple} & \textbf{CVAE} & \textbf{CTVAE} & N/P & $p$\\
		\hline
		\multicolumn{1}{c|}{\multirow{3}{*}{\emph{Fluency}}}
        & \emph{P1} &32.8(4.0) & -- & 40.4(6.0) & -- & 26.8(5.3) & $\mathbf{>0.05}$\\
		& \emph{P2} & -- & 25.6(1.6) & 32.4(1.7) & -- & 42.0(3.0) & $\mathbf{>0.05}$\\
		& \emph{P3} & -- & -- & 23.6(3.4) & \textbf{41.2}(2.9) & 35.2(6.0) & $<0.001$\\
		\hline
		\multicolumn{1}{c|}{\multirow{3}{*}{\emph{Topic relevance}}} & \emph{P1} &17.2(1.7) & -- & \textbf{63.2}(3.2) & --  & 19.6(2.9) & $<0.001$\\
		& \emph{P2} & -- & 28.0(1.9) & 34.4(1.1) & --  & 37.6(2.5) & $\mathbf{>0.05}$\\
		& \emph{P3} & -- & -- & 29.6(2.4) & \textbf{40.8}(1.9) & 29.6(4.1) & $<0.05$\\
		\hline
		\multicolumn{1}{c|}{\multirow{3}{*}{\emph{Informativeness}}} & \emph{P1} & 6.4(0.9) & -- & \textbf{79.6}(3.1) & -- & 14.0(2.7) & $<0.001$\\
		& \emph{P2} & -- & 28.4(2.3) & 28.0(2.1) & -- & 43.6(4.2) & $\mathbf{>0.05}$\\
		& \emph{P3} & -- & -- & 32.4(2.2) & \textbf{47.2}(1.1) & 20.4(2.6) & $<0.01$\\
		\hline
	\end{tabular}
    \caption{Average preference scores (std.) ($\%$) on fluency, topic relevance, and informativeness score between three model pairs (P1-P3), where N/P stands for "no preference" and $p$ denotes the $p$-value of a $t$-test between two models.}
	\label{tab:preference}
\end{table*}
\subsubsection{Diversity}
The percentages of distinct unigrams and bigrams \cite{li2015diversity} in the  generated top-5 responses were used to evaluate the diversity of the generated responses.
These two percentages, denoted as \emph{distinct-1} and \emph{distinct-2}, respectively,  represented the diversity at the n-gram level.
We also counted the percentage of unique response sentences, which evaluated the diversity of responses at sentence level.
The results for the four models are presented in Table \ref{tab:object_diversity}.
It can be found that the Seq2Seq model had the worst diversity at both the n-gram level and the sentence level. CVAE performed slightly better than CVAE-simple. And both CVAE models achieved better diversity at the n-gram level than that of the CTVAE model, especially for \emph{distinct-2}. According to the results of Section \ref{fluency_results}, the CVAE models performed worst on fluency performance, which may lead to higher diversity at the surface-text level. 
For the diversity at sentence level, the percentage of unique responses achieved by CTVAE was close to that of CVAE models.

\subsection{Subjective Evaluation}
It is difficult to evaluate the final performance of the generated conversation responses using objective metrics, such as BLEU.
It has been argued that such objective metrics for machine translation are very weakly correlated  with human judgment in dialog generation \cite{liu2016not}.
To evaluate the responses generated by our models in a more comprehensive and convincing manner, several groups of subjective ABX preference tests were conducted.
We randomly chose 50 posts from the test set and generated the top-5 responses from each model for each post.
The responses generated by two models were compared in each test.
Five native Chinese speakers with rich Sina Weibo experience were recruited for the evaluation.
For each test post, a pairs of top-5 responses generated by two models were presented in random order.
The evaluators were asked to judge which top-5 responses in each pair were preferred or if there was no preference on three subjective metrics: \emph{fluency}, \emph{topic relevance}, and \emph{informativeness} score.
{{Fluency}} was used to evaluate the quality of grammar and the semantic logic of responses.
{{Topic relevance}} measured whether a response matched the topic of the given post.
{{Informativeness}} measured how informative and interesting a response was.
In addition to calculating the average preference scores, the $p$-value of a $t$-test was adopted to measure the significance of the difference between two models.
Several significance levels were examined, including $p<0.05$, $p<0.01$, and $p<0.001$. $p>0.05$ indicated that there was no significant difference between two models.
The subjective evaluation results are presented in Table \ref{tab:preference}.

According to results of model pair P2 in Table \ref{tab:preference}, we can see that there is no significant difference on all three metrics between CVAE and CVAE-simple whose prior latent distribution was condition-independent, which implies that the effects of the condition-dependent prior distribution in the CVAE model were limited.
From the model pair P1, it can be found that CVAE outperformed Seq2Seq significantly on all metrics except fluency.
On the other hand, the results of model pair P3 show that CTVAE outperformed CVAE significantly on all three metrics, which confirms the effectiveness of our proposed CTVAE model.
These results indicate that our CTVAE model can derive $\mathbf{z}$ samples with better condition-dependency than CVAE model.
\begin{figure}[!t]
	\centering
	\includegraphics[width=3.5in]{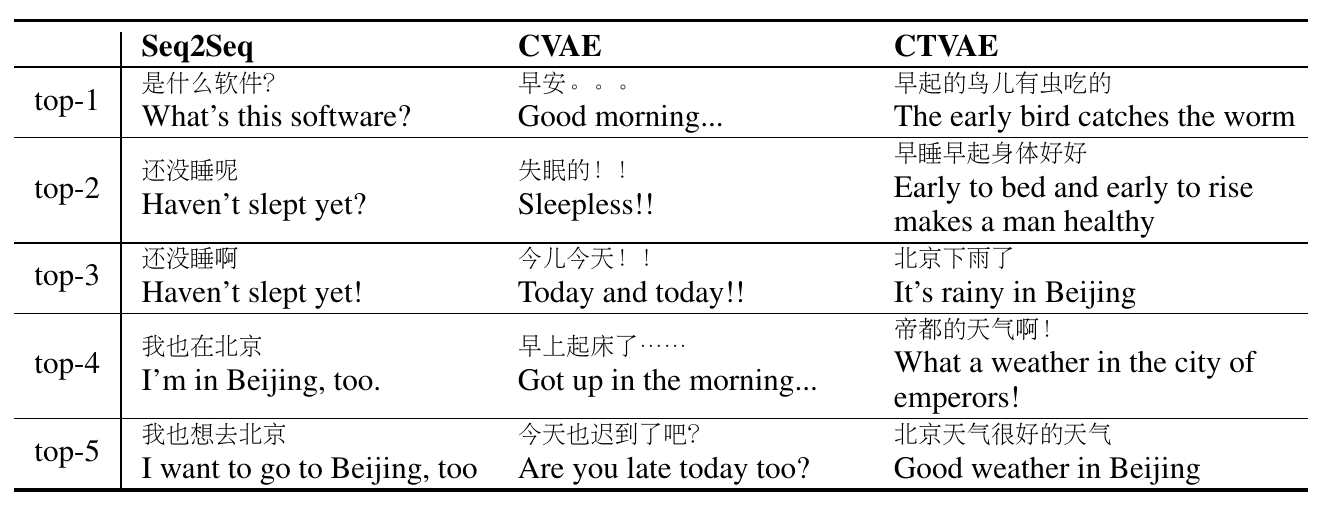}
	\caption{An example of the top-5 responses generated by Seq2Seq, CVAE, and CTVAE for the post ``It will be sunny after the rain in Beijing today. night owls, please get up early today.(\scriptsize{今天北京雨后天晴，熬夜人啊，早起吧。})".}\label{fig:case_study}
\end{figure}

\textbf{Case study} Figure \ref{fig:case_study} shows one typical example of the top-5 responses generated by the Seq2Seq, CVAE, and CTVAE models.
We can see that the Seq2Seq model tend to generate dull and generic responses.
Furthermore, the responses of CTVAE tended to be more topic relevant and informative than those of CVAE.

\section{Conclusion}
We have proposed a model named condition-transforming variational autoencoder (CTVAE) for diverse text generation.
In this model, the samples of latent  variable $\mathbf{z}$ are derived by performing a non-linear transformation on the combination of the input condition and the samples from a prior Gaussian distribution $\mathcal{N}(\mathbf{0}, \mathbf{I})$.
In our experiments on single-turn short text conversation, the CTVAE outperformed the Seq2Seq and CVAE models in both objective and subjective evaluations, which indicates that the CTVAE can derive $\mathbf{z}$ samples with better condition-dependency than CVAE models.
Applying the proposed CTVAE model to multi-turn conversation response generation and pursuing controllable sampling of the latent variable $\mathbf{z}$ will be our future work.

\vfill\pagebreak


\bibliographystyle{IEEEbib}
\bibliography{paper}

\begin{thebibliography}{10}

\bibitem{bahdanau2014neural}
Dzmitry Bahdanau, Kyunghyun Cho, and Yoshua Bengio,
\newblock ``Neural machine translation by jointly learning to align and
  translate,''
\newblock {\em arXiv preprint arXiv:1409.0473}, 2014.

\bibitem{DBLP:conf/emnlp/RushCW15}
Alexander~M Rush, Sumit Chopra, and Jason Weston,
\newblock ``A neural attention model for abstractive sentence summarization,''
\newblock in {\em Proceedings of the 2015 Conference on Empirical Methods in
  Natural Language Processing}, 2015, pp. 379--389.

\bibitem{DBLP:conf/nips/SutskeverVL14}
Ilya Sutskever, Oriol Vinyals, and Quoc~V Le,
\newblock ``Sequence to sequence learning with neural networks,''
\newblock in {\em Advances in neural information processing systems}, 2014, pp.
  3104--3112.

\bibitem{DBLP:conf/acl/ShangLL15}
Lifeng Shang, Zhengdong Lu, and Hang Li,
\newblock ``Neural responding machine for short-text conversation,''
\newblock in {\em Proceedings of the 53rd Annual Meeting of the Association for
  Computational Linguistics and the 7th International Joint Conference on
  Natural Language Processing (Volume 1: Long Papers)}, 2015, vol.~1, pp.
  1577--1586.

\bibitem{DBLP:conf/aaai/SerbanSBCP16}
Iulian~Vlad Serban, Alessandro Sordoni, Yoshua Bengio, Aaron~C. Courville, and
  Joelle Pineau,
\newblock ``Building end-to-end dialogue systems using generative hierarchical
  neural network models,''
\newblock in {\em Proceedings of the Thirtieth {AAAI} Conference on Artificial
  Intelligence, February 12-17, 2016, Phoenix, Arizona, {USA.}}, 2016, pp.
  3776--3784.

\bibitem{li2015diversity}
Jiwei Li, Michel Galley, Chris Brockett, Jianfeng Gao, and Bill Dolan,
\newblock ``A diversity-promoting objective function for neural conversation
  models,''
\newblock {\em arXiv preprint arXiv:1510.03055}, 2015.

\bibitem{xing2017topic}
Chen Xing, Wei Wu, Yu~Wu, Jie Liu, Yalou Huang, Ming Zhou, and Wei-Ying Ma,
\newblock ``Topic aware neural response generation.,''
\newblock in {\em AAAI}, 2017, pp. 3351--3357.

\bibitem{ghazvininejad2017knowledge}
Marjan Ghazvininejad, Chris Brockett, Ming-Wei Chang, Bill Dolan, Jianfeng Gao,
  Wen-tau Yih, and Michel Galley,
\newblock ``A knowledge-grounded neural conversation model,''
\newblock {\em arXiv preprint arXiv:1702.01932}, 2017.

\bibitem{wu2017neural}
Yu~Wu, Wei Wu, Dejian Yang, Can Xu, Zhoujun Li, and Ming Zhou,
\newblock ``Neural response generation with dynamic vocabularies,''
\newblock {\em arXiv preprint arXiv:1711.11191}, 2017.

\bibitem{li2016deep}
Jiwei Li, Will Monroe, Alan Ritter, Dan Jurafsky, Michel Galley, and Jianfeng
  Gao,
\newblock ``Deep reinforcement learning for dialogue generation,''
\newblock in {\em Proceedings of the 2016 Conference on Empirical Methods in
  Natural Language Processing}, 2016, pp. 1192--1202.

\bibitem{zhou2017mechanism}
Ganbin Zhou, Ping Luo, Rongyu Cao, Fen Lin, Bo~Chen, and Qing He,
\newblock ``Mechanism-aware neural machine for dialogue response generation.,''
\newblock in {\em AAAI}, 2017, pp. 3400--3407.

\bibitem{sohn2015learning}
Kihyuk Sohn, Honglak Lee, and Xinchen Yan,
\newblock ``Learning structured output representation using deep conditional
  generative models,''
\newblock in {\em Advances in Neural Information Processing Systems}, 2015, pp.
  3483--3491.

\bibitem{yan2016attribute2image}
Xinchen Yan, Jimei Yang, Kihyuk Sohn, and Honglak Lee,
\newblock ``Attribute2image: Conditional image generation from visual
  attributes,''
\newblock in {\em European Conference on Computer Vision}. Springer, 2016, pp.
  776--791.

\bibitem{zhao2017learning}
Tiancheng Zhao, Ran Zhao, and Maxine Eskenazi,
\newblock ``Learning discourse-level diversity for neural dialog models using
  conditional variational autoencoders,''
\newblock in {\em Proceedings of the 55th Annual Meeting of the Association for
  Computational Linguistics (Volume 1: Long Papers)}, 2017, vol.~1, pp.
  654--664.

\bibitem{serban2017hierarchical}
Iulian~Vlad Serban, Alessandro Sordoni, Ryan Lowe, Laurent Charlin, Joelle
  Pineau, Aaron~C Courville, and Yoshua Bengio,
\newblock ``A hierarchical latent variable encoder-decoder model for generating
  dialogues.,''
\newblock in {\em AAAI}, 2017, pp. 3295--3301.

\bibitem{yang2017generating}
Xiaopeng Yang, Xiaowen Lin, Shunda Suo, and Ming Li,
\newblock ``Generating thematic chinese poetry with conditional variational
  autoencoder,''
\newblock {\em arXiv preprint arXiv:1711.07632}, 2017.

\bibitem{kingma2014semi}
Diederik~P Kingma, Shakir Mohamed, Danilo~Jimenez Rezende, and Max Welling,
\newblock ``Semi-supervised learning with deep generative models,''
\newblock in {\em Advances in Neural Information Processing Systems}, 2014, pp.
  3581--3589.

\bibitem{kingma2013auto}
Diederik~P Kingma and Max Welling,
\newblock ``Auto-encoding variational bayes,''
\newblock {\em arXiv preprint arXiv:1312.6114}, 2013.

\bibitem{chen2017enhanced}
Qian Chen, Xiaodan Zhu, Zhen-Hua Ling, Si~Wei, Hui Jiang, and Diana Inkpen,
\newblock ``Enhanced lstm for natural language inference,''
\newblock in {\em Proceedings of the 55th Annual Meeting of the Association for
  Computational Linguistics (Volume 1: Long Papers)}, 2017, vol.~1, pp.
  1657--1668.

\bibitem{shang2016overview}
Lifeng Shang, Tetsuya Sakai, Zhengdong Lu, Hang Li, Ryuichiro Higashinaka, and
  Yusuke Miyao,
\newblock ``Overview of the ntcir-12 short text conversation task.,''
\newblock in {\em NTCIR}, 2016.

\bibitem{luong2015effective}
Thang Luong, Hieu Pham, and Christopher~D Manning,
\newblock ``Effective approaches to attention-based neural machine
  translation,''
\newblock in {\em Proceedings of the 2015 Conference on Empirical Methods in
  Natural Language Processing}, 2015, pp. 1412--1421.

\bibitem{kingma2014adam}
Diederik Kingma and Jimmy Ba,
\newblock ``Adam: A method for stochastic optimization,''
\newblock {\em arXiv preprint arXiv:1412.6980}, 2014.

\bibitem{bowman2016generating}
Samuel~R Bowman, Luke Vilnis, Oriol Vinyals, Andrew Dai, Rafal Jozefowicz, and
  Samy Bengio,
\newblock ``Generating sentences from a continuous space,''
\newblock in {\em Proceedings of The 20th SIGNLL Conference on Computational
  Natural Language Learning}, 2016, pp. 10--21.

\bibitem{mikolov2010recurrent}
Tom{\'a}{\v{s}} Mikolov, Martin Karafi{\'a}t, Luk{\'a}{\v{s}} Burget, Jan
  {\v{C}}ernock{\`y}, and Sanjeev Khudanpur,
\newblock ``Recurrent neural network based language model,''
\newblock in {\em Eleventh Annual Conference of the International Speech
  Communication Association}, 2010.

\bibitem{liu2016not}
Chia-Wei Liu, Ryan Lowe, Iulian Serban, Mike Noseworthy, Laurent Charlin, and
  Joelle Pineau,
\newblock ``How not to evaluate your dialogue system: An empirical study of
  unsupervised evaluation metrics for dialogue response generation,''
\newblock in {\em Proceedings of the 2016 Conference on Empirical Methods in
  Natural Language Processing}, 2016, pp. 2122--2132.

\end{thebibliography}
\end{CJK}
\end{document}